\documentclass[conference]{IEEEtran}
\usepackage{cite}
\usepackage{amsmath, amssymb, amsfonts}
\usepackage{algorithmic}
\usepackage{graphicx}
\usepackage{textcomp}
\usepackage{xcolor}
\usepackage{multicol, multirow}
\usepackage{adjustbox}
\usepackage{tabularx}
\usepackage{array}
\usepackage{float}
\def\BibTeX{{\rm B\kern-.05em{\sc i\kern-.025em b}\kern-.08em
    T\kern-.1667em\lower.7ex\hbox{E}\kern-.125emX}}

\begin{document}

\title{SpaRTAN: Spatial Reinforcement Token-based Aggregation Network for Visual Recognition}

\author{
    \IEEEauthorblockN{Quan Bi Pay\IEEEauthorrefmark{1}, Vishnu Monn Baskaran\IEEEauthorrefmark{1}, Junn Yong Loo\IEEEauthorrefmark{1}, KokSheik Wong\IEEEauthorrefmark{1} and Simon See\IEEEauthorrefmark{2}}
    \IEEEauthorblockA{\IEEEauthorrefmark{1}School of Information Technology, Monash University Malaysia}
    \IEEEauthorblockA{\IEEEauthorrefmark{2}NVIDIA AI Technology Center}
    \IEEEauthorblockA{\{quan.pay, vishnu.monn, loo.junnyong, wong.koksheik\}@monash.edu, ssee@nvidia.com}
}

\maketitle

\begin{abstract}
    The resurgence of convolutional neural networks (CNNs) in visual recognition tasks, exemplified by ConvNeXt, has demonstrated their capability to rival transformer-based architectures through advanced training methodologies and ViT-inspired design principles. However, both CNNs and transformers exhibit a simplicity bias, favoring straightforward features over complex structural representations. Furthermore, modern CNNs often integrate MLP-like blocks akin to those in transformers, but these blocks suffer from significant information redundancies, necessitating high expansion ratios to sustain competitive performance. To address these limitations, we propose SpaRTAN, a lightweight architectural design that enhances spatial and channel-wise information processing. SpaRTAN employs kernels with varying receptive fields, controlled by kernel size and dilation factor, to capture discriminative multi-order spatial features effectively. A wave-based channel aggregation module further modulates and reinforces pixel interactions, mitigating channel-wise redundancies. Combining the two modules, the proposed network can efficiently gather and dynamically contextualize discriminative features. Experimental results in ImageNet and COCO demonstrate that SpaRTAN achieves remarkable parameter efficiency while maintaining competitive performance. In particular, on the ImageNet-1k benchmark, SpaRTAN achieves 77. 7\% accuracy with only 3.8M parameters and approximately 1.0 GFLOPs, demonstrating its ability to deliver strong performance through an efficient design. On the COCO benchmark, it achieves 50.0\% AP, surpassing the previous benchmark by 1.2\% with only 21.5M parameters. The code is publicly available at [https://github.com/henry-pay/SpaRTAN].
\end{abstract}

\section{Introduction}\label{sec:intro}
Since the introduction of AlexNet~\cite{alexnet}, Convolutional Neural Networks (CNNs) have initiated a shift from hand-crafted feature engineering to data-driven feature extraction using end-to-end learning in visual recognition. The shift is governed by the property of translation equivariance, which is embedded in the sliding kernel operation. This introduces local inductive bias in CNNs, enabling feature recognition across different input resolutions. Motivated by the success of AlexNet, researchers have begun to explore various CNN-based architectural designs such as VGGNet~\cite{vggnet}, ResNet~\cite{resnet} and EfficientNet~\cite{efficientnet}. These recognition models, with a pyramidal network structure, can aggregate local responses with large effective receptive fields at various scales to capture global contextual information. However, they often fail to capture long-range dependency and neglect the importance of explicit global context modeling.\par
In 2020s, Vision Transformer (ViT)~\cite{vit} emerged as a significant alternative to CNNs, demonstrating state-of-the-art performance on various tasks such as image classification\cite{imagenet} and object detection~\cite{mscoco}. The strength of ViT is widely attributed to its self-attention mechanism~\cite{attention}, which enables the model to capture long-range dependencies in visual data. By encoding spatial information through global pairwise interactions, ViT enhances the generalizability of the model, albeit at the cost of longer training times and a higher number of parameters. To address these challenges, locality priors and pyramidal hierarchical layouts were reintroduced in subsequent models~\cite{swin, deit}. However, the quadratic complexity of self-attention, primarily due to the softmax function, continues to constrain computational efficiency~\cite{performer} and limits its applications to high-resolution fine-grained scenarios~\cite{swin-video}.\par
Inspired by ViT, ConvNeXt~\cite{convnext} has sparked the revival of CNNs, achieving compatible performance with ViT through advanced training techniques. Most modern CNNs~\cite{31x31, 51x51, conv2former} utilize a large kernel to capture long-range dependencies together with a ViT-style framework to remain competitive against ViT-based architectures. On the other hand, several works~\cite{hornet, moganet} have focused on introducing high-order spatial interactions into CNNs as a replacement for the self-attention mechanism. In short, feature extraction is refined in a local-global fashion by explicitly modeling global contextual information with a large kernel at the expense of computational cost. Further exploration of the ViT-based framework suggests a simple form of architectural design using pure Multi-Layer Perceptron (MLP) architecture with token-mixing and channel-mixing blocks~\cite{mlp-mixer, resmlp, wave-mlp}. These models have a lightweight and efficient design, but performance is still inferior to that of modern CNNs and ViT-based transformers.\par
\begin{figure}[tb]
    \centering
    \includegraphics[width=\linewidth]{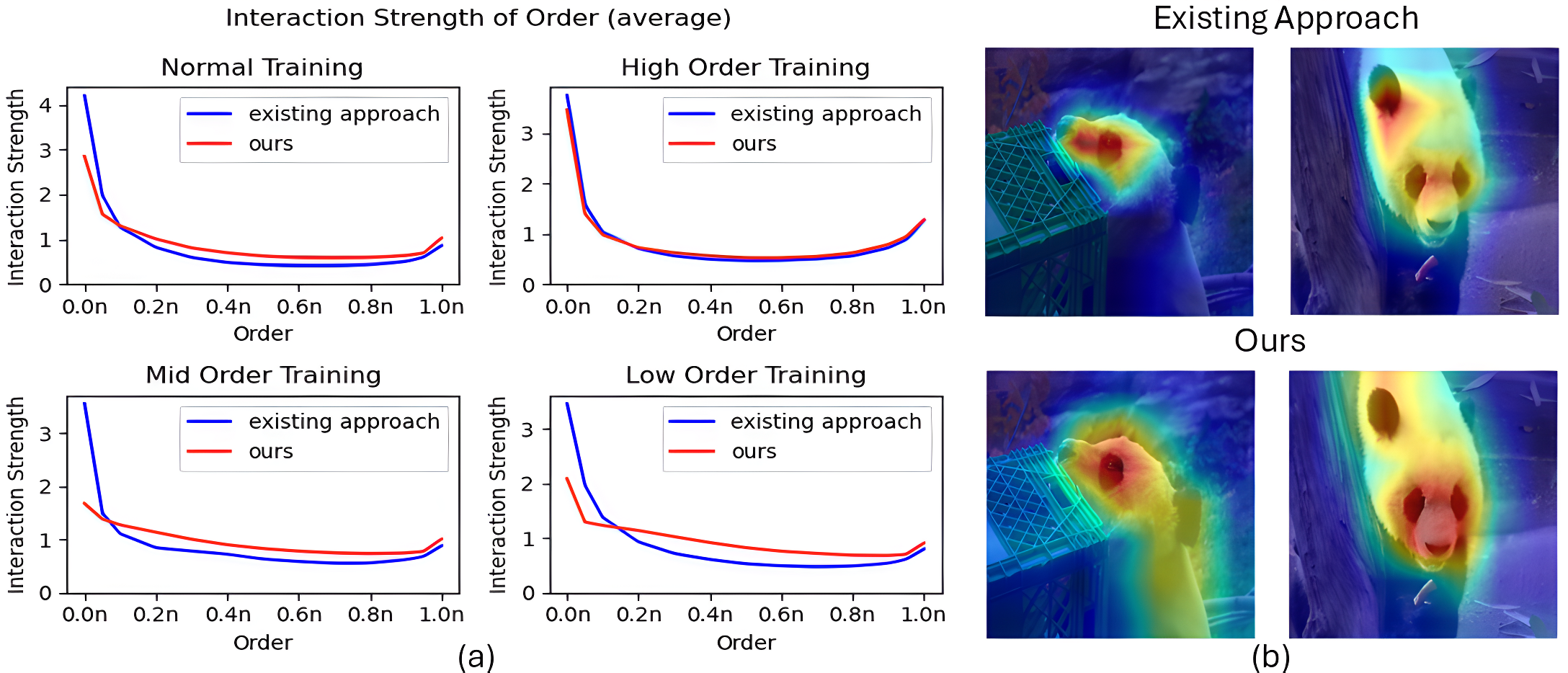}
    \caption{Line graphs in (a) illustrate the interaction strength at different orders of interaction between pixels for both the existing approach and the proposed method. The heatmaps in (b) present the impact of enhancing middle-order interactions. Notably, the proposed method successfully captures the full semantics of a panda, even in the presence of occlusions, whereas the existing method only identifies partial semantics, mainly focused around the eyes.}
    \label{fig:issues}
\end{figure}

Recent analysis from a game-theoretic perspective~\cite{game_theory_dnn} reveals that the capacity of modern CNNs to extract discriminative features is often undervalued, and hence they are not fully exploited for downstream tasks. Essentially, designs based on small kernels focus on simple, elementary visual concepts (low-order) that are generally shared across classes, while large kernels integrate global concepts (high-order), enabling comprehension of visual scenes using background elements. Nevertheless, complex textural and shape information (middle-order) that provides a discriminative understanding of patterns and structures is poorly harnessed~\cite{dnn_bottleneck}. Consequently, these models encode simple features that cannot clearly differentiate objects with similar contexts, which deteriorates their performance. Besides, these models attempt to encode complex interactions at the expense of redundancy and reduced efficiency. Not to mention, models that focus on high-order interactions are generally susceptible to adversarial attacks~\cite{dnn_bottleneck, game-adversarial}. Furthermore, a comparison between CNNs and transformer-based architecture~\cite{cnn_vs_transformer} highlights that both architectural designs favor simple features, which are often encoded in low-order or high-order interactions, and neglect discriminative structural information encoded in middle-order interactions. As illustrated in Fig.~\ref{fig:issues}, this highlights the representation bottleneck in CNN-based architectures and proposes a new direction in the design of efficient visual models.

\par
To this end, we propose a \textbf{Spa}tial \textbf{R}einforcement \textbf{T}oken-based \text{A}ggregated \textbf{N}etwork (SpaRTAN), which represents a lightweight architectural design that improves spatial- and channel-wise information processing. In detail, we conceive a convolution module that pursues discriminative visual representation learning through multi-order spatial interactions. Unlike previous works~\cite{hornet, moganet} that leverage large kernels to capture long-range dependency, stacked small kernels are utilized to achieve a similar effective receptive field with better computational efficiency. To refine and aggregate information across channels, a wave-based channel aggregation block is introduced to dynamically aggregate the information across channels according to their semantic context. Intuitively, each channel is formulated as a wave and modulated according to the selected maximally activated channel.

Extensive experiments on the ImageNet~\cite{imagenet} and COCO~\cite{mscoco} datasets demonstrate the consistent efficiency and competitive performance of the proposed architecture in image classification and object detection. SpaRTAN achieves \textbf{77.7\%} and \textbf{74.4\%} top-1 accuracy on ImageNet-1k with \textbf{3.8M} and \textbf{2.2M}. These results are achieved with a lower number of parameters and FLOPs. In terms of object detection, SpaRTAN shows a great performance gain, surpassing ResNet-18 and ResNet-34 by \textbf{3.6\%} and \textbf{1.1\%} respectively on the COCO dataset with reduced parameter space and processing cost. The results highlight the potential of the proposed network to maximize the utilization of the model parameters while maintaining competitive performance.

The contributions of this paper are as follows.
\begin{enumerate}
    \item \textbf{Spatial SMixer for Better Spatial Features Extraction:} We formulate a lightweight convolution module based on the concept of multi-order spatial interactions. This allows the SMixer to extract adaptive context across various scales and achieve full utilization of feature expressibility in convolutional kernels.
    \item \textbf{Wave-based CMixer for Dynamic Semantic Context Aggregation:} We introduce a wave-based channel aggregation module built on wave superposition. To the best of our knowledge, this is the first work to conceptualize individual channels as waves, enabling the reinforcement of pixel interaction strengths through maximally activated channels. This innovative approach effectively reduces inter-channel redundancy and enhances overall information flow.
    \item \textbf{A Lightweight Efficient Model with Competitive Results:} We combine the spatial SMixer and the wave-based CMixer to represent a new convolutional network. The network demonstrates a better balance between accuracy and computational efficiency in processing ImageNet and COCO datasets by capturing middle-order interactions between pixels.
\end{enumerate}
\section{Related Works}\label{sec:lit_rev}
\subsection{Vision Transformers}\label{sec:vit}
Self-attention mechanism~\cite{attention} is integrated into vision tasks by ViT~\cite{vit} through a simple patch embedding layer. However, to fully exploit the generalizability of the transformer, ViT is overly parameterized and requires large-scale training to achieve state-of-the-art performance. To resolve this issue, Swin~\cite{swin} re-introduces local inductive bias through shifted window multi-head self-attention into the ViT architecture. DeiT~\cite{deit}, on the other hand, focuses on training strategy and showcases that the ViT-based architecture can detect and recognize small-scale elements using knowledge distillation and advanced data augmentation techniques. However, quadratic complexity within the self-attention mechanism still persists and greatly impacts the scalability of the model. EfficientFormer~\cite{efficientformer} utilizes a dimension-consistent design with an efficient token-mixer mechanism to achieve a fully transformer-based efficient architecture. In contrast, MobileViT~\cite{mobilevit} opts for a hybrid design, using MobileNetV2~\cite{mobilenetv2} as the baseline infrastructure integrated with ViT blocks for explicit global context modeling. Building upon the hybrid design, ShuffleViTNet~\cite{shufflevitnet} uses depthwise and $1\times 1$ convolution to replace convolution operations in a ViT block for efficient design with minimal memory access cost. Nonetheless, these architectures still suffer performance issues, highlighting a sub-optimal accuracy-efficiency trade-off.\par
\subsection{Convolutional Neural Networks}\label{sec:cnn}
Recently, motivated by the results of ConvNeXt~\cite{convnext}, several methods have integrated large convolution kernels to achieve long-range dependencies in CNNs~\cite{31x31, 51x51, van, conv2former}. However, computational overhead incurred by large convolutional kernels affects model training and inference speed. To resolve this issue, efficient high-order spatial interactions are explored~\cite{hornet, moganet}. These designs comprises spatial mixing block, SMixer$(\cdot)$ and channel mixing block, CMixer$(\cdot)$ formulated as
\begin{align}
    Y &= X + \text{SMixer}(\text{Norm}(X))\label{eq:smixer}, \\
    Z &= Y + \text{CMixer}(\text{Norm}(Y))\label{eq:cmixer},
\end{align}
where Norm$(\cdot)$ refers to normalization layer and $X\in\mathbb{R}^{C\times H\times W}$ is an image. This design highly resembles that of a transformer to capture spatial information in $n$-order interactions, followed by a simple contextual aggregation mechanism. We show that by carefully designing the SMixer and CMixer modules, one can achieve a lightweight yet robust model design.\par
\subsection{Multi-Layer Perceptron}\label{sec:mlp}
Following~\eqref{eq:smixer} and~\eqref{eq:cmixer}, MLP-like architectures composed solely of linear layers with non-linear activation functions are explored due to their simple and efficient architectural design. MLP-Mixer~\cite{mlp-mixer} introduces a token mixing block to capture spatial information, functions as SMixer, and a channel mixing block to extract characteristics for each token, acting as CMixer. ResMLP~\cite{resmlp} introduces a new learnable affine transformation that can replace the normalization layer to stabilize the training of linear layers. The formulation offers explicit global contextual learning which is independent across channels. Subsequently, Wave-MLP~\cite{wave-mlp} proposes to view tokens as wave particles composed of amplitude and phase. This formulation enhances the visual representation of the network through dynamic information aggregation based on the semantic context. Despite the simple design, their performance is poor in comparison to CNNs and vision transformers.\par
\section{Methodology}\label{sec:method}
\subsection{Overview}\label{sec:method_overview}
\begin{figure*}[tb]
    \centering
    \includegraphics[width=0.85\linewidth]{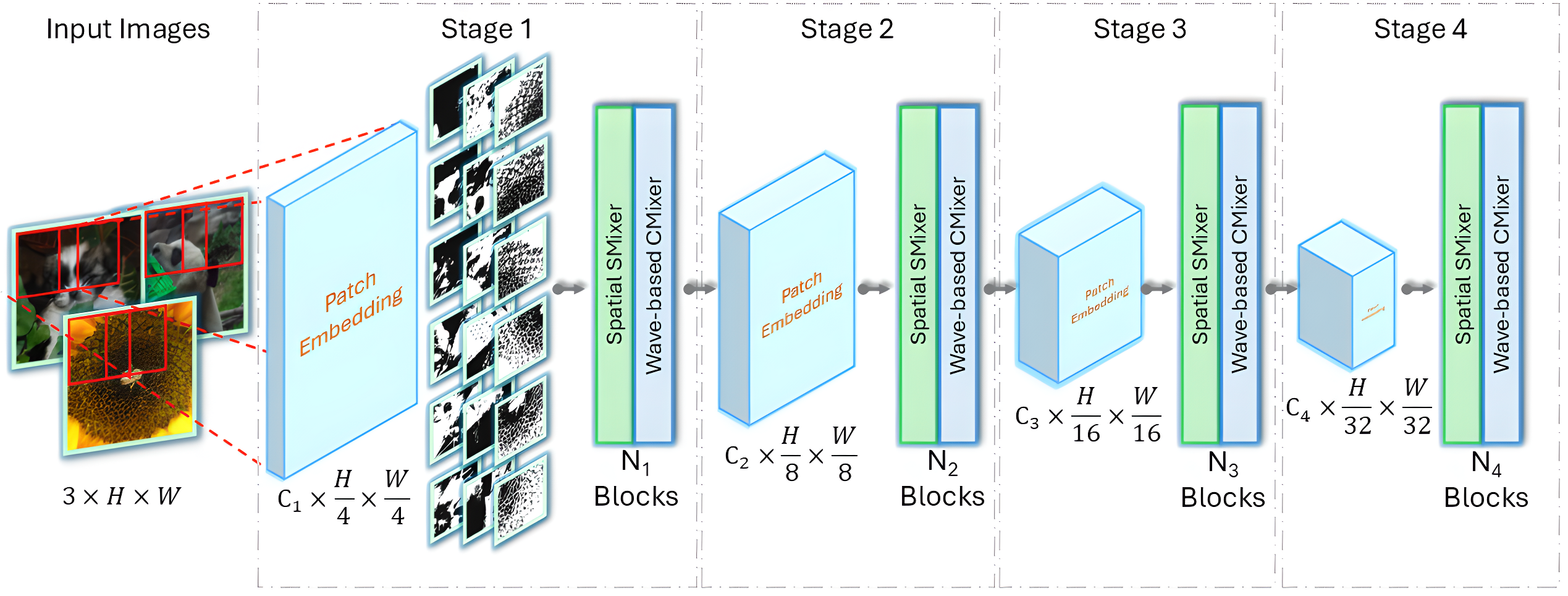}
    \caption{A high-level graphical illustration of SpaRTAN consisting spatial SMixer and wave-based CMixer. The proposed architecture utilizes a hierarchical design with four stages, each comprising a patch embedding layer followed by $N_i$ blocks of SMixer and CMixer.}
    \label{fig:architecture}
\end{figure*}

Building on modern CNNs~\cite{convnext, conv2former}, our model adopts a pyramidal structure with 4 stages, as shown in Fig.~\ref{fig:architecture}. This hierarchical design is achieved using a patch embedding module to downsample the resolution of feature maps by a factor of 2 at each stage except stage 1. In stage 1, an overlapping patch embedding is used, consisting of two $3\times 3$ convolutions with a stride of 2. A stacked convolution allows efficient initial feature extraction with a gradually expanded effective receptive field. Meanwhile, non-overlapping patch embedding, which is a $2\times 2$ convolution with a stride of 2, is used in other stages.

Subsequently, the embedded features are passed to the main building block, consisting of the SMixer and CMixer modules. SMixer is mainly responsible for feature extraction, typically implemented as sliding kernel operations such as convolution to capture local information or attention-based mechanisms to gather global contextual information. As discussed in Section~\ref{sec:intro}, such operations are prone to simplicity bias, often overlooking robust and more expressive interactions. To encourage the network in exploiting information in originally ignored interactions or features, we propose to leverage the idea of varying-sized kernels to attain adaptive feature extraction in low- and high-frequency regions. In contrast, 
an MLP structure with an expand ratio, $r$, is the standard CMixer function to aggregate contextual information across channels. An additional Squeeze-and-Excitation (SE) layer~\cite{senet} can be inserted to further refine the features. However, due to information redundancy across channels, a higher $r$ is expected to achieve competitive performance, resulting in additional computational and memory overheads. To mitigate this, we introduce a wave-based channel aggregation module to dynamically reallocate channel-wise features, achieving better parameter utilization. After the final stage, a global average pooling (GAP) layer and a linear layer are added for image classification.\par
\subsection{Spatial SMixer}\label{sec:conv-smixer}
The area of coverage of a convolution operation is determined by its kernel size, $k$. A small kernel covers a smaller area and thus can capture finer details and rapid changes more effectively. Such localized details correspond to high-frequency components where there are quick transitions, such as edges and textures. In contrast, a large kernel is capable of capturing global information, which stays relatively consistent throughout the image. This is often referred to as the low-frequency components, including background elements and large objects. It has been shown that the performance of architectures with only small kernel convolutions such as ResNet~\cite{resnet} and EfficientNet~\cite{efficientnet} falls short of a transformer-based architecture. On the other hand, architectures that pursue only large-kernel convolution~\cite{31x31, 51x51} fail to encode expressive features and are prone to simplicity bias, despite having competitive performance as transformers. To resolve such a dilemma, we propose composite convolutions of varying sizes as an instantiation of SMixer with input $X\in\mathbb{R}^{C\times H\times W}$ such that
\begin{equation}\label{eq:proposed-smixer}
    \text{SMixer}(X) = \mathcal{S}(\text{FD}(X)),
\end{equation}
where FD$(\cdot)$ is a feature decomposition module adapted from~\cite{moganet} and $\mathcal{S}(\cdot)$ is a convolutional module that extracts and aggregates both low- and high-frequency components to capture expressive features. In FD$(\cdot)$, there are two complementary counterparts, fine-grained local features extracted using $1\times 1$ convolution, and global contextual information retrieved using GAP. The reweighting scheme, $\gamma \odot \left(X - \text{GAP}(X)\right)$, in $FD(\cdot)$ improves the diversity of spatial characteristics, encouraging a more varied feature distribution that enforces inherently ignored interactions~\cite{moganet}. 

\begin{figure}[tb]
    \centering
    \includegraphics[width=\linewidth]{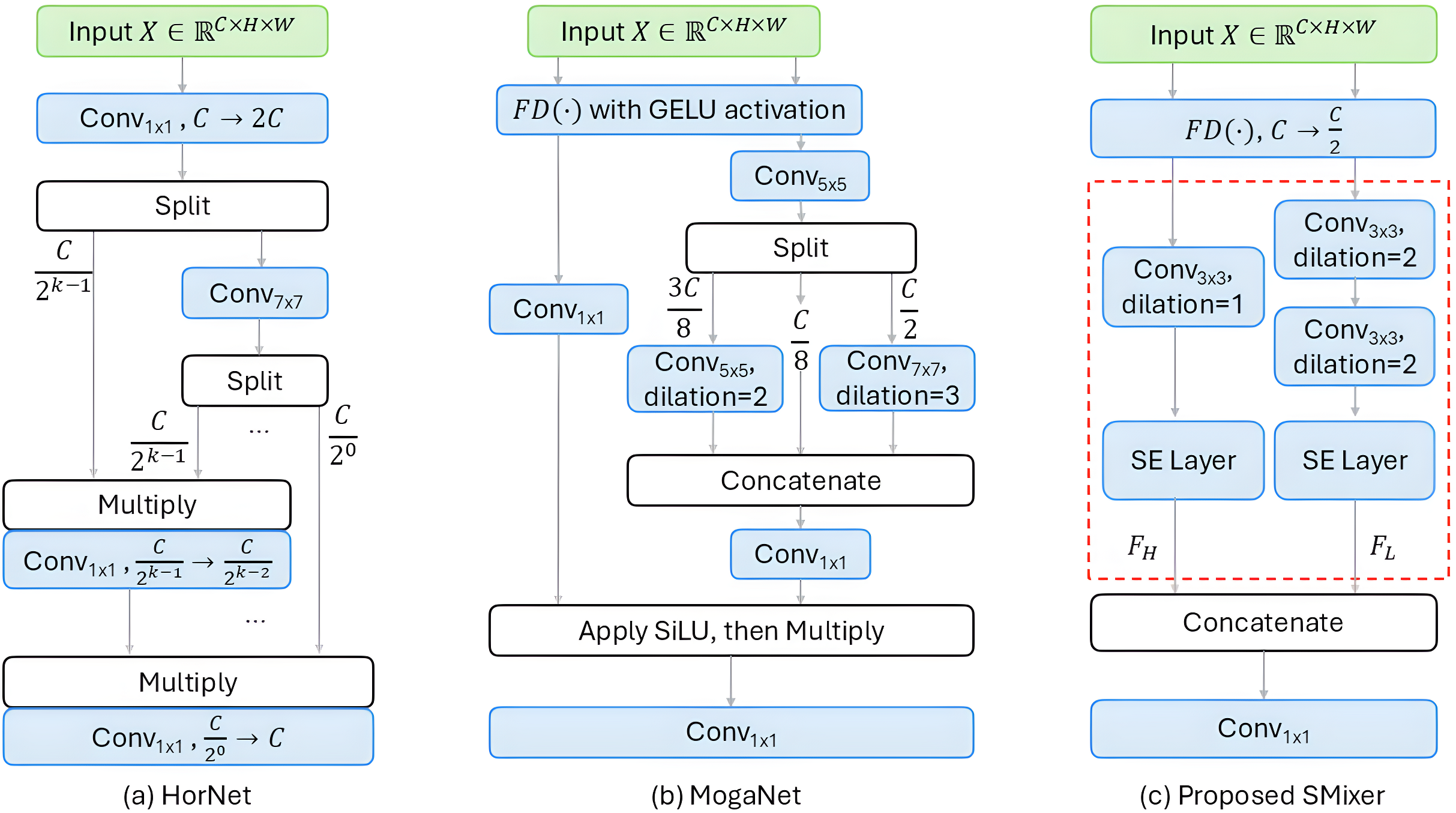}
    \caption{A detailed layout of various convolution-based high-order spatial interaction mechanisms for comparison. (a) is adapted from HorNet~\cite{hornet}, leveraging a recursive function to extract arbitrary n-order spatial interactions. On the other hand, (b) is adapted from MogaNet~\cite{moganet}, utilizing varying kernel sizes and dilation factors to achieve multi-order spatial interactions. The proposed method (c) employs a two-branch architecture to extract low- and high-frequency components with a convolution of varying receptive fields.}
    \label{fig:multi-order}
\end{figure}

Subsequently, convolutions of varying kernel size are ensembled to encode multi-order features classified into low- and high-frequency components. In previous work, multi-order feature extraction is achieved using recursive gated convolutions~\cite{hornet} or multi-order gated aggregation~\cite{moganet}, as shown in Fig.~\ref{fig:multi-order}(a) and (b). Both employ large kernel convolutions (up to $7\times 7$) to extract features at varying proximity. In HorNet, multi-order operation is achieved through recursive depthwise convolution on projected features with different numbers of channels. In contrast, MogaNet~\cite{moganet} utilizes a straightforward design which relies on large convolution kernels with various dilation factors to capture interactions between distant patches. Instead of working on the same set of feature maps split across channel dimension as in previous works, we leverage a 2-branch architecture to encode low- and high-frequency components with large and small kernels, respectively, as illustrated in Fig.~\ref{fig:multi-order}(c). Given the input $X\in\mathbb{R}^{C\times H\times W}$, Conv$_{3\times 3, \text{dilation}=1}$ and Conv$_{5\times 5, \text{dilation}=2}$ are applied to retrieve the high- and low-frequency components, $F_H$ and $F_L$, respectively. To achieve better efficiency, Conv$_{5\times 5, \text{dilation}=2}$ is replaced by a stacked Conv$_{3\times 3, \text{dilation}=2}$ following the design of~\cite{vggnet}. Note that convolutions share the same effective receptive field, but using stacked $3\times 3$ convolutions can save up to $6\%$ in GFlops without much degrading performance, as shown in Table~\ref{tab:kernel}.

Note that the complexity, $T$, and memory access cost, $M$, of convolution can be written as
\begin{align}
    T &= \underbrace{C_{\text{in}} \times C_{\text{out}}}_{\text{channel projection}}\times\underbrace{K\times K}_{\text{kernel size}}\times\underbrace{H\times W}_{\text{input size}}, \\
    M &= \underbrace{C_{\text{in}} \times C_{\text{out}} \times K \times K}_{\text{kernel weights}} + \underbrace{C_{\text{in}}\times H \times W}_{\text{input memory}} + \underbrace{C_{\text{out}}\times H\times W}_{\text{output memory}},
\end{align}
where $C_{\text{in}}$ is input channel, $C_{\text{out}}$ is output channel, $K$ is kernel size, $H$ and $W$ are the input resolutions. In the earlier stage, $HW\gg C_{\text{in}}C_{\text{out}}$ while in the later stage, $C_{\text{in}}C_{\text{out}} \gg HW$ due to hierarchical design. Simply replacing all convolutions with depthwise convolution as in previous works may not yield the optimal efficiency improvement. As pointed out in~\cite{efficientnet}, depthwise convolution incurs additional overhead at an early stage as it cannot fully utilize modern accelerators. Hence, at stages 1 and 2, a full convolution is utilized while depthwise convolution is employed at stages 3 and 4, giving a better accuracy-efficiency trade-off, as shown in Table~\ref{tab:convolution}.

To adaptively aggregate the extracted low- and high-frequency components, an additional SE layer~\cite{senet} is added after the convolutions. Through feature recalibration, discriminative multi-order feature representations are achieved by suppressing trivial interactions based on global contextual information. Taking the output from the convolutions, $\mathcal{S}(\cdot)$ in~\eqref{eq:proposed-smixer} can be instantiated as
\begin{equation}
    S = \text{Conv}_{1\times 1}\left(\left[\text{SE}(F_H);\text{SE}(F_L)\right]\right).
\end{equation}
The final projection processes the features in a spatially coherent manner, resulting in a more refined feature representation. This design better captures discriminative features without the cost-consuming attention operations.\par
\subsection{Wave-based CMixer}\label{sec:wave-cmixer}
\begin{figure}[tb]
    \centering
    \includegraphics[width=\linewidth]{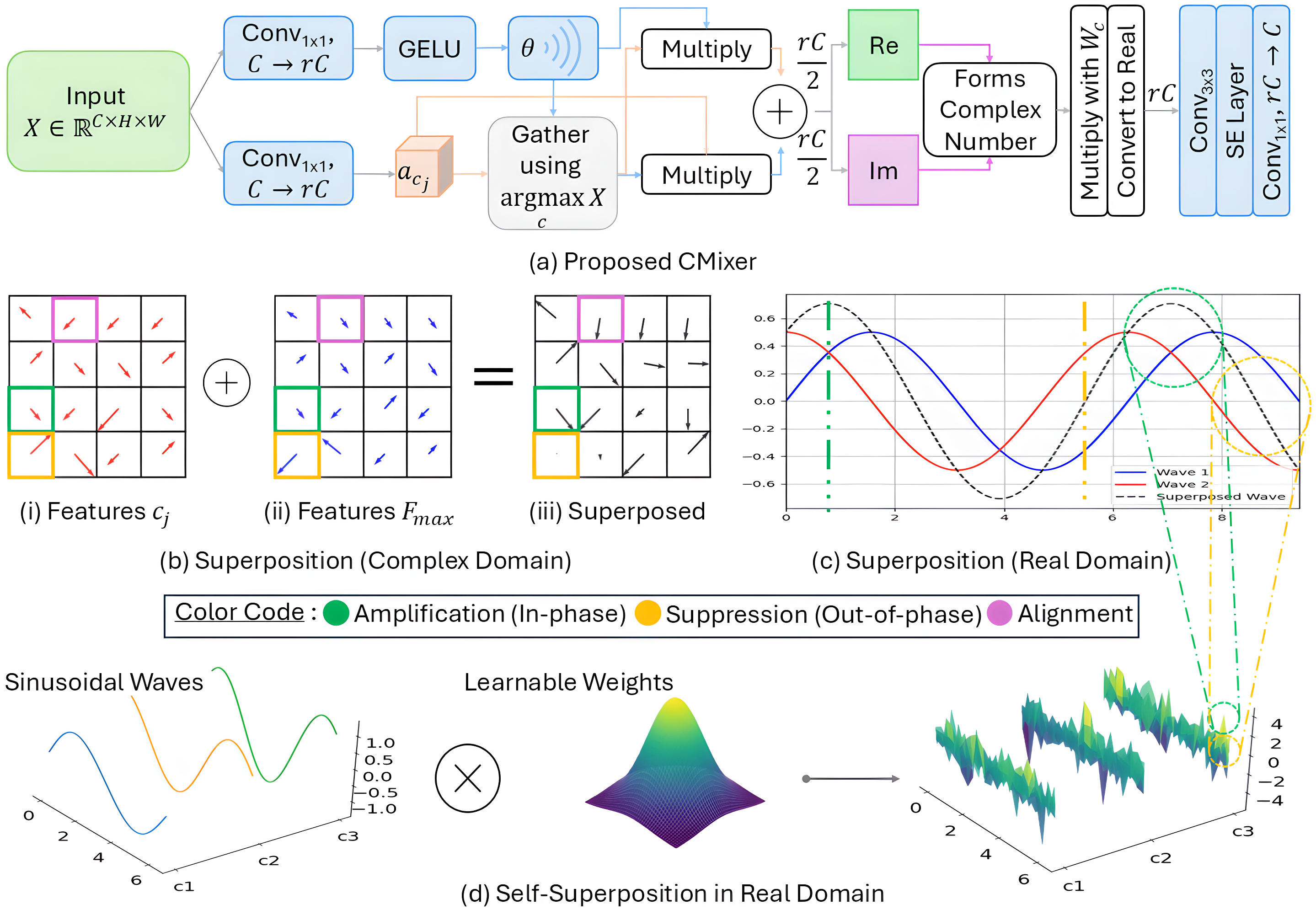}
    \caption{A detailed overview of wave-based CMixer is outlined in (a). Graphical illustrations in (b) and (c) showcase the interactions between the wave under superposition in complex and real domains. (d) highlights the effect of self-superposition in the real domain after being modulated by complex weights. The peak and trough correspond to enhanced wave superposition, while the x-intercept is the result of cancellation between opposing waves.}
    \label{fig:wave}
\end{figure}

As stated in Section~\ref{sec:method_overview}, vanilla MLP requires an additional number of parameters using a high expand ratio $r$, typically set to 4 or 8, to achieve competitive performance. Moreover, their fixed weights often do not adapt to the varying semantic content across channels in different input images, thereby restricting the ability to form context-aware representations. To overcome this drawback, we propose a wave-based channel aggregation module, as illustrated in Fig.~\ref{fig:wave}(a), as an instantiation of CMixer with input $X\in\mathbb{R}^{C\times H\times W}$ such that
\begin{equation}
    \text{CMixer} = \mathcal{C}(\mathcal{W}(X)),
\end{equation}
where $\mathcal{W}(\cdot)$ is wave-based channel aggregation block and $\mathcal{C}(\cdot)$ is feature refinement module.

Inspired by Wave-MLP~\cite{wave-mlp}, we first construct $\mathcal{W}(\cdot)$ by treating the feature channels as an oscillating wave consisting of the amplitude $a_{c_j}$ and the phase $\theta_j$. The feature channels are denoted as $C=[c_1,c_2,\dots,c_n]$ where $c_j\in\mathbb{R}^{H\times W}$. Following Euler's formula, the phase information can be disentangled into $\cos$ and $\sin$ functions in a complex domain. Hence, the feature channel can be expressed as a complex number, i.e.
\begin{equation}\label{eq:wave}
    c_j = |a_{c_j}|\odot e^{i\theta_j} = |a_{c_j}|\left(\cos\theta_j+i\sin\theta_j\right),
\end{equation}
where $i$ represents the imaginary unit. Note that the complex value in~\eqref{eq:wave} is represented as two real values, forming the real and imaginary parts. Since each channel is a real-valued component representing the oscillating wave, we simplify by treating half of the channels as a $\sin$ wave and the other half as a $\cos$ wave. Together, they form the phase information in~\eqref{eq:wave}.

To retain the most expressive features, we propose a superposition mechanism with $F_{\max}\in\mathbb{R}^{1\times H\times W}$ that contains the maximum value of $C$. As illustrated in Fig.~\ref{fig:wave}(b), this mechanism represents another form of similarity measure where the features relevant to $F_{\max}$ will be amplified while the trivial features will be suppressed. In addition, the resultant vectors are more aligned with $F_{\max}$, encouraging a sharp focus on $F_{\max}$ for dynamic channel aggregation. From a real domain perspective, the interaction between the channels is greatly influenced by phase information, as described in Fig.~\ref{fig:wave}(c). Intuitively, the features are enhanced when they are in phase and suppressed when they are completely out of phase. This allows dynamic channel aggregation based on the key semantic context, measured by the raw values of the feature maps.

To refine the wave information, a complex weight, $W_c\in\mathbb{C}^{\frac{C}{2}\times 1\times 1}$, is introduced to modulate channels in the complex domain. Suppose $w_c = a + bi$, the complex multiplication will initiate superposition on both the real and imaginary part, i.e.
\begin{equation}\label{eq:self-superposition}
    |a_{c_j}|\left(a\cos\theta_j - b\sin\theta_j\right) + i|a_{c_j}|\left(a\sin\theta_j + b\cos\theta_j\right),
\end{equation}
where the negative sign can be absorbed into the phase information using the identity $-\sin\theta_j = \sin -\theta_j$. Essentially, the introduction of complex weights leads to a self-superposition operation in which the real and imaginary components modulate each other for the refinement of the features, as shown in~\eqref{eq:self-superposition} and Fig.~\ref{fig:wave}(d). Modulation generates waves with varying frequencies, improving interaction capture between channels. Observe that the wave formulation indicates a shift from the spatial domain to the frequency domain. Hence, such modulation is motivated by the equivalence between multiplication in the frequency domain and global circular convolution in the spatial domain~\cite{gfn}. This captures both short- and long-term interactions by adjusting the learnable weights $W_c$, which in theory cover the entire frequency spectrum.

Following~\cite{wave-mlp}, both the amplitude and phase information is extracted using linear projections of the input $X$ with modulus operation absorbed into the phase. However, the oscillatory nature of sinusoidal functions often leads to frequent direction switching during weight updates, resulting in unstable training. To mitigate this issue, we opt for a linear approximation of the sinusoidal waves using a point convolution with a non-linear activation function. Together with learning in the complex domain, the proposed mechanism sets a clear distinction with~\cite{wave-mlp}. This concludes the wave-based aggregation block.

To initiate an inverse operation from the frequency domain to the spatial domain while maintaining as much information as possible, $\mathcal{C}(\cdot)$ with input $X\in\mathbb{R}^{C\times H\times W}$ is instantiated as
\begin{equation}
    \mathcal{C}(X) = \text{Conv}_{1\times 1}(\text{SE}(\text{Conv}_{3\times 3}(X))).
\end{equation}
Similar to SMixer, a mixture of full convolution and depthwise convolution is used at different stages to ensure an optimal accuracy-efficiency trade-off. The results in Table~\ref{tab:arch-ablation} verify the effectiveness of wave-based CMixer compared to vanilla MLP and MLP with an SE module in achieving competitive representation ability under a small $r$. This suggests that a wave-based formulation improves parameter utilization while reducing information redundancies across the channels.\par
\subsection{Implementation Details}
For efficient lightweight design, we construct SpaRTAN for 2 model sizes (SpaRTAN-XT and SpaRTAN-T) with different numbers of blocks and channels at each stage. Table~\ref{tab:arch_implementation} details the model configurations. GELU and BatchNorm form the normalization and activation function after the convolution layer. However, LayerNorm is used in~\eqref{eq:smixer} and~\eqref{eq:cmixer}. SiLU is applied instead in the patch embedding layer. This combination is empirically determined as described in Table~\ref{tab:activation} and Table~\ref{tab:normalization}.

\begin{table}[tb]
    \centering
    \caption{Architectural Configurations}
    \label{tab:arch_implementation}
    \begin{tabular}{c|c|c|c|c}
        \hline
         Stage & Output Size & Layer Setting & SpaRTAN-XT & SpaRTAN-T \\
         \hline
         \multirow{5}{*}{S1} & \multirow{5}{*}{$\frac{H\times W}{4}$} & \multirow{2}{*}{Patch Embed} & \multicolumn{2}{c}{Conv$_{3\times 3}$, stride 2, $C / 2$} \\
         & & & \multicolumn{2}{c}{Conv$_3{\times 3}$, stride 2, $C$} \\
         \cline{3-5}
         & & Channels & \multicolumn{2}{c}{32} \\
         \cline{3-5}
         & & \# Blocks & \multicolumn{2}{c}{3} \\
         \cline{3-5}
         & & Expand Ratio & \multicolumn{2}{c}{4} \\
         \hline
         \multirow{4}{*}{S2} & \multirow{4}{*}{$\frac{H\times W}{8}$} & Patch Embed & \multicolumn{2}{c}{Conv$_{2\times 2}$, stride 2} \\
         \cline{3-5}
         & & Channels & \multicolumn{2}{c}{64} \\
         \cline{3-5}
         & & \# Blocks & \multicolumn{2}{c}{3} \\
         \cline{3-5}
         & & Expand Ratio & \multicolumn{2}{c}{4} \\
         \hline
         \multirow{4}{*}{S3} & \multirow{4}{*}{$\frac{H\times W}{16}$} & Patch Embed & \multicolumn{2}{c}{Conv$_{2\times 2}$, stride 2} \\
         \cline{3-5}
         & & Channels & 96 & 128 \\
         \cline{3-5}
         & & \# Blocks & 10 & 12 \\
         \cline{3-5}
         & & Expand Ratio & \multicolumn{2}{c}{2} \\
         \hline
         \multirow{4}{*}{S4} & \multirow{4}{*}{$\frac{H\times W}{32}$} & Patch Embed & \multicolumn{2}{c}{Conv$_{2\times 2}$, stride 2} \\
         \cline{3-5}
         & & Channels & 192 & 256 \\
         \cline{3-5}
         & & \# Blocks & \multicolumn{2}{c}{2}\\
         \cline{3-5}
         & & Expand Ratio & \multicolumn{2}{c}{2} \\
         \hline
    \end{tabular}
\end{table}
\section{Experiments}\label{sec:exp}
In this section, we present the result of experiments on popular vision tasks such as image classification and object detection to examine and compare the proposed architecture with the leading lightweight network architectures. The experiments are implemented with PyTorch and run on NVIDIA A100 GPUs.

\subsection{ImageNet Classification}\label{sec:imagenet-evaluation}
\subsubsection{Settings}
To assess the performance of the proposed architecture, we performed experiments using the ImageNet-1k~\cite{imagenet} dataset, a benchmark for classification tasks. The dataset comprises 1,000 categories, with about 1.2 million images in the training set and 50,000 images in the evaluation set. Unless stated otherwise, the input image resolution is set to $224 \times 224$ for training. All models are trained for 300 epochs using the AdamW optimizer, with a training setup that includes a batch size of 2048, a base learning rate of $2.5 \times 10^{-3}$, a weight decay of 0.03, and a cosine learning rate scheduler featuring a 20-epoch warm-up phase. We employ a comprehensive set of data augmentation and regularization techniques to enhance performance. These include Random Resized Crop, Horizontal Flip, RandAugment (with a magnitude of 7), Mixup ($\alpha = 0.2$), CutMix, Random Erasing, and Stochastic Depth.

\subsubsection{Quantitative Result}
As shown in Table~\ref{tab:imagenet_result}, SpaRTAN-T achieves competitive performance compared to state-of-the-art architectures while significantly optimizing the number of parameters and FLOPs. Remarkably, SpaRTAN-T achieves a top-1 accuracy of $77.1\%$ with only 3.8M parameters and 0.83 GFLOPs, outperforming models that require more than 4M parameters. Furthermore, the smaller variant, SpaRTAN-XT, delivers competitive results with only 2.2M parameters, offering an even more resource-efficient solution. These results underscore that CNN architectures employing small kernels can rival or surpass the efficiency and effectiveness of models utilizing attention mechanisms and large-kernel convolutions. We attribute this improvement to the spatial SMixer and wave-based CMixer, which together enable more effective parameter utilization and yield a superior accuracy-efficiency trade-off.

\begin{table}[tb]
    \centering
    \caption{ImageNet-1K Classification Results}
    \label{tab:imagenet_result}
    \begin{tabular}{c|ccc|c}
        \hline
         \multirow{2}{*}{Model} & \multirow{2}{*}{Image Size} & Param & FLOPs & {Top-1 Acc.} \\
         & & (M) & (G) &  (\%) \\
         \hline
         ResNet-18~\cite{resnet} & 224\textsuperscript{2} & 11.7 & 1.80 & 71.5 \\
         DeiT-T~\cite{deit} & 224\textsuperscript{2} & 5.7 & 1.08 & 74.1 \\
         PVT-T~\cite{pvt} & 224\textsuperscript{2} & 13.2 & 1.60 & 75.1 \\
         ViT-C~\cite{vit-c} & 224\textsuperscript{2} & 4.6 & 1.10 & 75.3 \\
         VAN-B0~\cite{van} & 224\textsuperscript{2} & 4.1 & 0.88 & 75.4 \\
         EfficientFormerv2~\cite{efficientformerv2} & 224\textsuperscript{2} & 3.6 & 0.42 & 75.7 \\
         MobileOne-S1~\cite{mobileone} & 224\textsuperscript{2} & 4.8 & 0.83 & 75.9 \\
         EfficientNet-B0~\cite{efficientnet} & 224\textsuperscript{2} & 5.3 & \textbf{0.39} & 77.1 \\
         MogaNet-XT~\cite{moganet
         } & 256\textsuperscript{2} & 3.0 & 1.04 & 77.2 \\
         Swin-1G~\cite{swin} & 224\textsuperscript{2} & 7.3 & 1.00 & 77.3 \\
         ConvNeXt-XT~\cite{convnext} & 224\textsuperscript{2} & 7.4 & 0.60 & 77.5 \\
         \hline
         SpaRTAN-XT & 224\textsuperscript{2} & \textbf{2.2} & 0.64 & 73.9 \\
         SpaRTAN-XT & 256\textsuperscript{2} & \textbf{2.2} & 0.83 & 74.4 \\
         SpaRTAN-T & 224\textsuperscript{2} & 3.8 & 0.83 & 77.1 \\
         SpaRTAN-T & 256\textsuperscript{2} & 3.8 & 1.08 & \textbf{77.7} \\
         \hline
    \end{tabular}
\end{table}

\begin{figure}[tb]
    \centering
    \includegraphics[width=\linewidth]{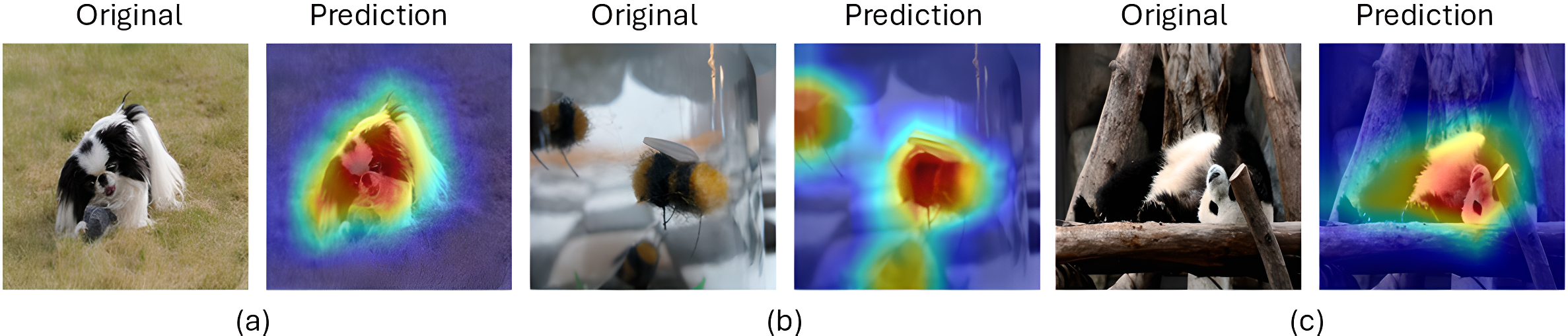}
    \caption{Grad-CAM analysis on (a) Japanese Spaniel, (b) Bees and (c) Giant Panda. The activation maps showcase the ability of SpaRTAN in identifying the complete semantics of the objects, even in the presence of occlusions.}
    \label{fig:single-object}
\end{figure}

\begin{table*}[tb]
    \centering
    \caption{COCO 2017 Object Detection Results}
    \label{tab:coco_result}
    \begin{tabular}{cc|ccc|cccccc}
        \hline
        \multirow{2}{*}{Architecture} & \multirow{2}{*}{Backbone} & \multirow{2}{*}{Epochs} & Params & FLOPs
        & \multirow{2}{*}{AP\textsuperscript{val}} & \multirow{2}{*}{AP$^{\text{val}}_{50}$} & \multirow{2}{*}{AP$^{\text{val}}_{75}$} & \multirow{2}{*}{AP$^{\text{val}}_{S}$} & \multirow{2}{*}{AP$^{\text{val}}_{M}$} & \multirow{2}{*}{AP$^{\text{val}}_{L}$}  \\
         & & & (M) & (G)
         & & & & & & \\
         \hline
         Deformable-DETR~\cite{deformable-detr} & ResNet-50 & 50 & 40.0 & 173.00 & 46.2 & 65.2 & 50.0 & 28.8 & 49.2 & 61.7 \\
         RT-DETR~\cite{rt-detr} & ResNet-18 & 72 & 20.2 & \textbf{61.20}
         & 46.4 & 63.7 & 50.3 & 28.4 & 49.7 & 63.0 \\
         DAB-Deformable-DETR~\cite{dab-detr} & ResNet-50 & 50 & 48.0 & 195.00 & 46.9 & 66.0 & 50.8 & 30.1 & 50.4 & 62.5 \\
         DN-Deformable-DETR~\cite{dn-detr} & ResNet-50 & 50 & 48.0 & 195.00 & 48.6 & 67.4 & 52.7 & 31.0 & 52.0 & 63.7 \\
         RT-DETR~\cite{rt-detr} & ResNet-34 & 72 & 31.4 & 93.30
         & 48.8 & 66.7 & 52.6 & 30.5 & 52.2 & 66.0 \\
         \hline
         RT-DETR~\cite{rt-detr} & SpaRTAN-XT (Ours) & 70 & \textbf{20.0} & 72.80
         & 48.5 & 66.1 & 52.6 & 30.1 & 51.9 & 66.0 \\
         RT-DETR~\cite{rt-detr} & SpaRTAN-T (Ours) & 70 & 21.5 & 75.90
         & \textbf{50.0} & \textbf{68.1} & \textbf{54.0} & \textbf{32.7} & \textbf{53.8} & \textbf{68.0} \\
         \hline
    \end{tabular}
\end{table*}

\subsubsection{Qualitative Result}
\begin{figure}[t]
    \centering
    \includegraphics[width=\linewidth]{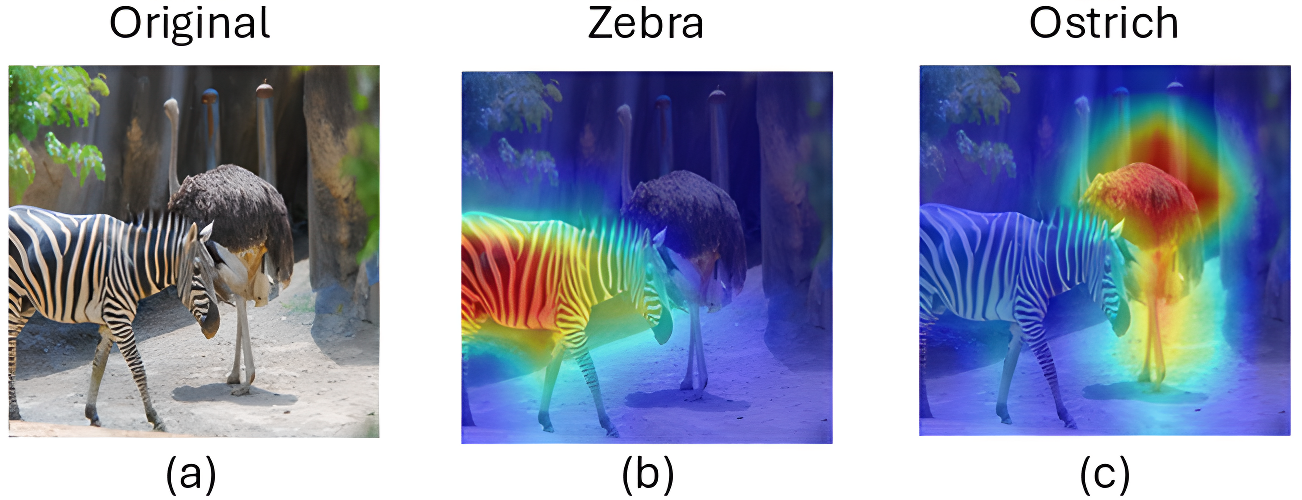}
    \caption{Activation maps on the image with multiple distinct objects. SpaRTAN can correctly recognize semantics with respect to each object.}
    \label{fig:multi-object}
\end{figure}

To better comprehend the advantages of the proposed architecture, we visualize the activation maps generated using Grad-CAM~\cite{gradcam}. As shown in Fig.~\ref{fig:single-object}(a), our model effectively captures all the semantic information from the Japanese Spaniel while maintaining a sharp focus on facial structure. Additionally, Figs.~\ref{fig:single-object}(b) and (c) demonstrate the model's ability to recognize multiple objects sharing the same semantic characteristics and accurately extract semantic features even in the presence of occlusions. These showcase the capability of the model to retain a high level of accuracy and clarity in recognizing the objects' key features. We extended our evaluation to include images that contain multiple types of objects, as illustrated in Fig.~\ref{fig:multi-object}. In these scenarios, the model demonstrated its ability to accurately distinguish and interpret the semantics of different objects, such as identifying and differentiating between a zebra and an ostrich. This highlights the effectiveness of the model in handling complex scenes with diverse elements.

\par
\subsection{COCO 2017 Object Detection}\label{sec:coco-evaluation}
\subsubsection{Settings}
We evaluate the proposed architecture in the object detection task using the COCO~\cite{mscoco} dataset, a highly utilized benchmark. The dataset contains 80 categories across 200,000 images. The experiment utilizes RT-DETR~\cite{rt-detr}, a real-time object detector based on DETR, as the baseline architecture. We replicate the exact training setup from~\cite{rt-detr}, including data augmentation and Exponential Moving Average (EMA). Training is conducted on the \textit{COCO train2017} dataset, with evaluation performed on \textit{COCO eval2017}.

\begin{table}[tb]
    \centering
    \caption{Ablation on SMixer and CMixer}
    \label{tab:arch-ablation}
    \begin{tabular}{c|c}
        \hline
        Module & Top-1 Acc. (\%) \\
        \hline
        Baseline (Convolution + MLP) & 71.9 \\
        \hline
        SMixer + MLP &  72.4 \\
        SMixer + MLP + SE & 72.6 \\
        \hline
        Convolution + MLP + SE & 72.3 \\
        Convolution + CMixer & 72.8 \\
        \hline
        SMixer + CMixer & \textbf{73.9}
         \\
        \hline
    \end{tabular}
\end{table}

\subsubsection{Quantitative Result}
Table~\ref{tab:coco_result} shows that object detectors using the proposed architecture as a backbone achieve better results with improved efficiency. Notably, SpaRTAN-T variant outperforms all other models with more than 40 million parameters, demonstrating its superior efficiency and effectiveness despite having nearly 2 times smaller parameters. It is important to note that SpaRTAN-T is much more efficient than the ResNet-34 RT-DETR variant and outperforms it by $1.2\%$ using 10M less parameters. The smaller variant of the model, built using SpaRTAN-XT, outperforms the ResNet-18 RT-DETR variant with a similar number of parameters and achieves competitive results compared to the ResNet-34 variant. Note that the increase in 1.5M parameters from the SpaRTAN-XT to the SpaRTAN-T variant results in a $1.5\%$ performance boost, whereas a nearly 10M parameter increase in the ResNet variant only leads to a $2.4\%$ performance improvement. This indicates that our backbone architecture can utilize the parameters more effectively to provide rich semantic features than the widely used ResNet architecture.\par
\subsection{Ablation Studies}\label{sec:ablation}
\subsubsection{SMixer and CMixer}
Taking a non-linear projection using $3\times 3$ convolutions as the SMixer and a vanilla MLP as the CMixer, we construct a baseline model to evaluate the effectiveness of the proposed SMixer and CMixer. Table~\ref{tab:arch-ablation} details the performance of each variation. Each proposed module contributes to improving the overall performance, indicating its enhanced capability over the baseline modules. In particular, SMixer and CMixer individually boost accuracy by $0.5\%$ and $0.9\%$, respectively, over the baseline model. Pairing the SMixer with the SE module yields a modest performance gain of $0.3\%$, over the baseline using the SE module alone. However, combining the SMixer with the CMixer results in a more significant improvement of $2\%$. Meanwhile, the effectiveness of replacing the $5\times 5$ kernel with two $3\times 3$ kernels is evaluated in Table~\ref{tab:kernel}. With that, the number of FLOPs and the number of parameters are reduced by $\approx 6\%$ and $\approx 1.4\%$ respectively with an acceptable performance drop.

\begin{table}[tb]
    \centering
    \caption{Ablation on Kernel Size in SMixer}
    \label{tab:kernel}
    \begin{tabular}{c|cc|c}
        \hline
        Kernel Size & Params (M) & FLOPs (G) & Top-1 Acc. (\%)\\
        \hline
        3 & \textbf{2.07} & \textbf{0.64} & 73.90 \\
        5 & 2.10 & 0.68 & \textbf{73.93} \\
        \hline
    \end{tabular}
\end{table}

\begin{table}[t]
    \centering
    \caption{Ablation on Activation Function in Patch Embedding Layer and Building Blocks (SMixer+CMixer)}
    \label{tab:activation}
    \begin{tabular}{c|cccc}
        \hline
        Activation (Patch Embedding) & GELU & GELU & SILU & SILU\\
        Activation (Building Block) & GELU & SILU & GELU & SILU \\
        \hline
        Top-1 Acc. (\%) & 72.8 & 73.2 & \textbf{73.9} & 72.9 \\
        \hline
    \end{tabular}
\end{table}

\subsubsection{Activation and Normalization}
We conducted an ablation of the activation function used in the patch embedding layer and the building blocks, SMixer and CMixer. The results in Table~\ref{tab:activation} indicate that the gating effect of SiLU performs best in patch embedding layers, while the training-friendly GELU function is more suitable to be used with SMixer and CMixer. We also examine the effectiveness of normalization layers by ablating the types of normalization applied after the convolution layer and before SMixer in~\eqref{eq:smixer} and CMixer in~\eqref{eq:cmixer}. For simplicity, we compare the two most widely used normalization techniques, Batch Normalization and Layer Normalization. As illustrated in Table~\ref{tab:normalization}, Batch Normalization performs better than Layer Normalization when used after convolution. Layer Normalization, on the other hand, works well when it is used before SMixer and CMixer.

\begin{table}[tb]
    \centering
    \caption{Ablation on Normalization Layer After Convolution and Before SMixer+CMixer}
    \label{tab:normalization}
    \begin{tabular}{c|c|c}
        \hline
        Normalization & Normalization & Top-1 Acc. \\
        (After Convolution) & (Before SMixer and CMixer) & (\%) \\
        \hline
        BatchNorm & BatchNorm & 73.1 \\
        BatchNorm & LayerNorm & \textbf{73.9} \\
        LayerNorm & BatchNorm & 72.6 \\
        LayerNorm & LayerNorm & 72.2 \\
        \hline
    \end{tabular}
\end{table}

\subsubsection{Type of Convolution}
We ablate the convolution types (full [F], depthwise [D], and hybrid [H]) in SMixer and CMixer to evaluate the accuracy-efficiency trade-off. In the hybrid setting, half of the stage uses F, and the other half uses D. Using a batch size of 1024 to evaluate throughput, the results in Table~\ref{tab:convolution} show that the hybrid setting achieves the best balance. Although depthwise convolution has fewer parameters and FLOPs compared to full convolution, its throughput suffers because it cannot fully utilize modern accelerators. However, using full convolution in all stages significantly increases the number of parameters and FLOPs, and while it achieves the highest accuracy, it also results in reduced throughput.

\begin{table}[tb]
    \centering
    \caption{Ablation Study on Types of Convolution}
    \label{tab:convolution}
    \begin{tabular}{c|cccc|c}
        \hline
        \multirow{2}{*}{Types}
        &  Params & FLOPs & MACs & Throughput & Top-1 Acc.\\
         & (M) & (G) & (M) & (img/sec) & (\%) \\
         \hline
         F & 3.3 & 0.78 & 754.2 & $3008\pm 6.3$ & \textbf{74.1} \\
         D & \textbf{2.2} & \textbf{0.57} & \textbf{539.2} & $3640\pm 17.3$ & 73.2 \\
         H & \textbf{2.2} & 0.64 & 614.4 & $\mathbf{3730\pm 8.3}$ & 73.9 \\
         \hline
    \end{tabular}
\end{table}
\section{Conclusion}\label{sec:conclusion}
This paper proposes SpaRTAN, a modern CNN-based visual recognition architecture. Built upon modern high-order spatial interaction modules for extracting discriminative spatial features, we present a simple spatial SMixer, leveraging convolutions with varying receptive fields to extract features tied to middle-order interactions. This is followed by a wave-based CMixer to aggregate contextual information across channels effectively using a smaller expand ratio $r$ compared to vanilla MLP. Experiments conducted with ImageNet and COCO datasets verify the proposed model's performance and efficiency. In conclusion, the proposed mechanism represents a good balance of performance and efficiency and is beneficial for various vision tasks. 

Nevertheless, several avenues remain open for investigation as part of future work. First, larger kernels could be explored for the proposed method with the aim of preserving state-of-the-art accuracy and efficiency. Additionally, to assess the generalizability of the proposed model, future work could include experiments on a broader range of downstream tasks, such as semantic segmentation and pose estimation. Another possible future direction is to investigate the behavior of SpaRTAN under common model compression strategies, including pruning and low-bit quantization, which could offer further insights into SpaRTAN’s efficacy in resource-constrained environments.
\section*{Acknowledgment}

This work was supported in part by the Advanced Computing Platform at Monash University Malaysia. The authors would like to thank the anonymous reviewers for their constructive comments and feedback.

\bibliography{section/ref}

\end{document}